\documentclass[10pt,twocolumn,letterpaper]{article}

\usepackage[pagenumbers]{cvpr} 

\definecolor{cvprblue}{rgb}{0.21,0.49,0.74}
\usepackage[pagebackref,breaklinks,colorlinks,allcolors=cvprblue]{hyperref}
\usepackage{multirow}
\usepackage{enumitem}
\usepackage{svg}
\usepackage{xcolor}
\usepackage[ruled,vlined]{algorithm2e}
\usepackage[table]{xcolor}

\definecolor{darkgreen}{RGB}{0,110,0}
\definecolor{bestcolor}{RGB}{255,245,200}   
\definecolor{secondcolor}{RGB}{220,235,255} 

\SetCommentSty{mycommentstyle}
\def\confName{CVPR}
\def\confYear{2027}

\title{FlexReader: Towards Adaptive Decoding Paths for Efficient Arbitrary-Length Scene Text Recognition}

\author{
\shortstack[c]{
Zhibin Ma$^{1,2}$,
Pengwen Dai$^{1,2}$,
Zhigang Yao$^{3}$,
Xugong Qin$^{4}$,\\
Wenqi Ren$^{1}$,
Xiaochun Cao$^{1,2}$\\[6pt]
$^{1}$Shenzhen Campus of Sun Yat-sen University\\[2pt]
$^{2}$Shenzhen Key Laboratory of Adversarial Artificial Intelligence\\[2pt]
$^{3}$Chongqing University \qquad
$^{4}$Nanjing University of Science and Technology
}
}

\begin{document}
\maketitle
\begin{abstract}
Existing Scene Text Recognition (STR) models are inherently coupled with a specific decoding strategy by their training paradigms, limiting their flexibility in deployment across diverse recognition scenarios. In this paper, we propose \textbf{\textit{FlexReader}}, an STR model that supports multiple decoding strides, enabling a single model to achieve an optimal accuracy–efficiency trade-off across different datasets. Specifically, we first introduce a Sub-string Perception module to perceive the target sub-string from visual features, followed by multiple parallel decoding blocks that independently decode each character in the sub-string. During inference, FlexReader supports decoding with arbitrary strides within the perceived sub-string range. Furthermore, we propose Adaptive Prefix Decoding, which adaptively selects the decoding stride at each step based on the model’s confidence in the decoded characters, thereby identifying the optimal decoding path for each input.
Experimental results demonstrate that FlexReader exhibits robust recognition capability across a diverse range of benchmarks, including those featuring long text, curved text, and occlusions. It consistently outperforms existing decoding strategies in recognition accuracy while achieving up to 6× decoding speedup over autoregressive decoding.

\end{abstract}    
\section{Introduction}
Scene Text Recognition (STR) aims to extract textual information from natural scene images. It serves as a fundamental component of visual understanding and has broad applications in autonomous driving, embodied intelligence, augmented reality, and other fields.

Unlike conventional Optical Character Recognition~\cite{arxiv/got/HRWei, arxiv/monkeyocrv1.5/JRZhang, cvpr26/ppocrv5/CCui} scenarios, STR entails recognizing text under challenging visual conditions, such as complex backgrounds and irregular layouts, while posing stringent requirements on inference efficiency. However, existing methods struggle to balance recognition accuracy and inference efficiency. For example, AR methods~\cite{tpami19/aster/BGShi, cvpr20/seed/ZQiao, eccv22/parseq/Bautista, iccv23/maerec/QJiang} achieve high recognition accuracy but incur substantial decoding overhead, while parallel decoding~\cite{cvpr20/srn/DLYu, eccv22/matrn/Na, tpami25/cppd/YKDu} and CTC-based~\cite{tpami17/crnn/BGShi, ijcai22/svtr/YKDu, arxiv/svtrv2/YKDu} methods enable efficient inference at the expense of limited character dependency modeling. Although recent mask diffusion-based decoding approaches~\cite{mm21/pimnet/ZQiao, ijcv25/ipad/XMYang, aaai26/mdiff/YKDu} effectively balance accuracy and efficiency, their fixed-length design limits their ability to handle texts with varying lengths. Moreover, existing methods are typically optimized with a single objective (e.g. the AR loss), which restricts them to a single decoding strategy at inference time. However, the difficulty of scene text varies considerably across real-world scenarios, necessitating adaptive decoding strategies that can dynamically adjust to the input difficulty for an optimal accuracy–efficiency trade-off. 

\begin{figure}[t]
    \centering
    \includegraphics[width=0.47\textwidth]{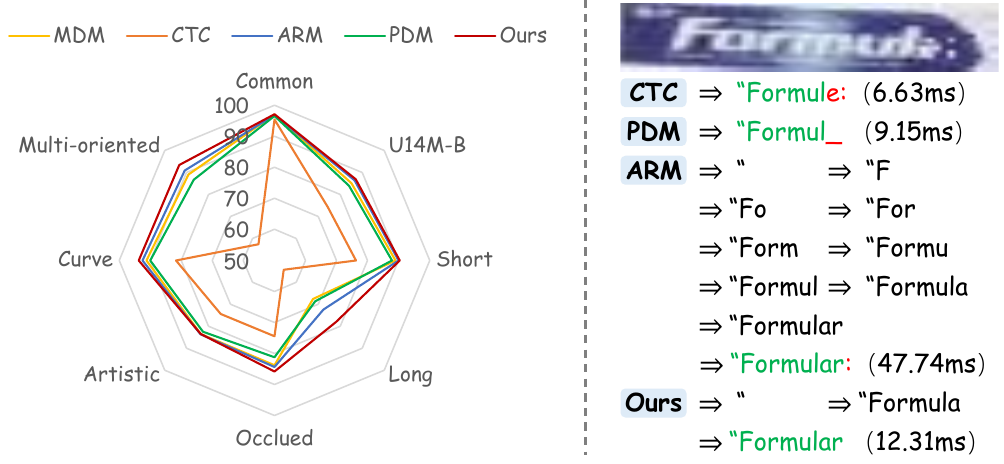}
    \caption{\textbf{Left:} Comparison of different decoding methods.\textbf{Right:} A representative recognition example, where our method adaptively selects the optimal decoding path.
} 
    \label{fig1}
\end{figure}

Inspired by the success of multi-token prediction (MTP) in large language models (LLMs)~\cite{icml24/mtp/Gloeckle, nips25/lmtp/XHLiu, icml24/medusa/TLCai, eccv26/pmtp/LXiang}, we explore its application to STR decoding for flexible and efficient inference. However, unlike language modeling, where future tokens can be inferred from contextual representations, STR requires fine-grained alignment between each predicted character and its corresponding visual evidence. Therefore, how to develop a STR framework that can effectively exploit multi-token prediction to accelerate decoding without sacrificing character-level recognition accuracy across diverse text lengths remains an open challenge.

In this work, we propose \textbf{FlexReader}, an efficient and flexible STR network to address the aforementioned challenges. First, we decompose the decoding process into two components: \textit{Sub-string Perception} and \textit{Multi-branch Prediction}. Unlike the next-token perception adopted in AR decoding, Sub-string Perception captures the visual evidence of the next $L$ characters conditioned on the previously decoded character sequence, and encodes the entire sub-string into a single token (as illustrated in Fig.~\ref{fig2}(c)). Subsequently, Multi-branch Prediction utilizes $L$ parallel decoding branches to individually recover each character from the encoded sub-string. This decoding strategy decouples the perception and decoding processes, enabling flexible inference-time decoding with different decoding strides ranging from $1$ to $L$. Second, we propose \textit{Adaptive Prefix Decoding}, which enables the model to adaptively determine the decoding length for each forward pass based on prediction confidence, achieving an optimal speed–accuracy trade-off within a single model. Compared with the speculative decoding commonly used in LLMs~\cite{icml23/spedec/Leviathan, icml24/medusa/TLCai}, our decoding approach eliminates the complex draft prediction and verification process, enabling faster inference with a lightweight model. 

The proposed FlexReader provides two key advantages. (1) Flexbility: FlexReader allows a single model to decode with arbitrary step sizes and adaptively selects the decoding stride based on sample difficulty. This capability effectively mitigates the speed–accuracy dilemma faced by existing approaches. (2) Efficiency: Through Adaptive Prefix Decoding, FlexReader autonomously searches for an optimal decoding path by leveraging prediction confidence as prior guidance. Furthermore, sub-string perception enables the decoder to incorporate textual information from future sub-strings, breaking the limitation of conventional autoregressive decoding that relies solely on previously decoded tokens. The main contributions of this work are fourfold:

\begin{enumerate}[label=(\arabic*)]
    \item We pioneer the integration of multi-token prediction into the STR task, introducing a new decoding paradigm that enables efficient recognition of text sequences with a wide range of lengths.

    \item We decompose the decoding process into \textit{Sub-string Perception} and \textit{Multi-branch Prediction}, which enables fine-grained sub-string-level alignment and allows a single model to perform variable-step decoding.

    \item We propose \textit{Adaptive Prefix Decoding}, a dynamic decoding strategy that adaptively adjusts the number of predicted tokens for an optimal accuracy-efficiency trade-off.

    \item Extensive experiments demonstrate the effectiveness of the proposed \textbf{FlexReader}. It achieves state-of-the-art performance across diverse benchmarks while delivering a \textbf{4--6$\times$} speedup in decoding.
\end{enumerate}


\section{Related works}
With the rapid advancement of deep learning~\cite{nature2025/deeplearning/Lecun, cvpr16/KMHe/resnet}, numerous deep neural network-based approaches have been proposed for scene text recognition (STR). Based on their decoding mechanisms, these methods can be broadly categorized into two groups: CTC-based methods~\cite{arxiv/cnnctc/YZGao, bmvc16/starnet/WLiu, tpami17/crnn/BGShi, aaai20/WYHu/gtc} and encoder-decoder~\cite{aaai23/trocr/MHLi, aaai26/mdiff/YKDu, cvpr20/srn/DLYu, tpami19/aster/BGShi, pr19/moran/CJLuo} methods. 

\subsection{CTC-based methods}
Connectionist Temporal Classification (CTC) was originally proposed for weakly aligned speech recognition~\cite{icml06/ctc/Graves}, and CRNN~\cite{tpami17/crnn/BGShi} first introduced CTC into STR for visual feature–character alignment. Subsequent CTC-based methods have primarily focused on improving visual feature extraction, while the CTC decoding paradigm has remained unchanged. For example, SVTR~\cite{ijcai22/svtr/YKDu} introduced two types of blocks to capture intra- and inter-character patterns, while SVTRv2~\cite{arxiv/svtrv2/YKDu} incorporated dedicated modules to enhance spatial perception and semantic modeling. Despite their efficient decoding and simple training, these methods rely on one-dimensional feature sequences, limiting their ability to handle complex scene text, particularly irregular text.

\subsection{Encoder-decoder methods}
Encoder-decoder methods typically employ a visual encoder to extract visual features and a specially designed decoder to decode text from these features. Based on their decoding strategies, they can be broadly categorized into autoregressive models (ARMs) and parallel decoding models (PDMs). ARMs~\cite{tpami19/aster/BGShi, aaai26/one2seq/ZBMa, eccv20/autostr/HZhang, cvpr20/seed/ZQiao, eccv22/parseq/Bautista, iccv23/maerec/QJiang, eccv20/robustscanner/XYYue} perform sequential decoding, predicting one character at each step. This sequential decoding process enables effective utilization of semantic information, which benefits the recognition of challenging text under occlusion, blur, and low-light conditions. Consequently, ARMs are widely adopted in high-accuracy STR models. However, this multi-step decoding process significantly slows down inference. In contrast, PDMs~\cite{cvpr21/abinet/SCFang, tpami25/cppd/YKDu, eccv22/matrn/Na, eccv22/mgpstr/PWang} predict all characters in a single forward pass. Although this significantly improves decoding speed, their recognition performance still lags significantly behind that of ARMs. Mask Diffusion Models (MDMs)~\cite{mm21/pimnet/ZQiao, ijcv25/ipad/XMYang, aaai26/mdiff/YKDu} provide a compromise between the two paradigms. They adopt an easy-first strategy and iteratively decode and remask tokens for a fixed number of iterations. However, these methods require the text length to be predetermined before decoding, making them less flexible when handling extremely long or short text.

In this work, we propose FlexReader as an alternative to the aforementioned approaches. As a new exploration of encoder-decoder methods, FlexReader incorporates multi-token prediction into the STR decoder. Our approach brings greater flexibility and efficiency to STR decoding, as demonstrated in the remainder of this paper.
\begin{figure*}[t!]
    \centering
    \includegraphics[width=1.0\textwidth]{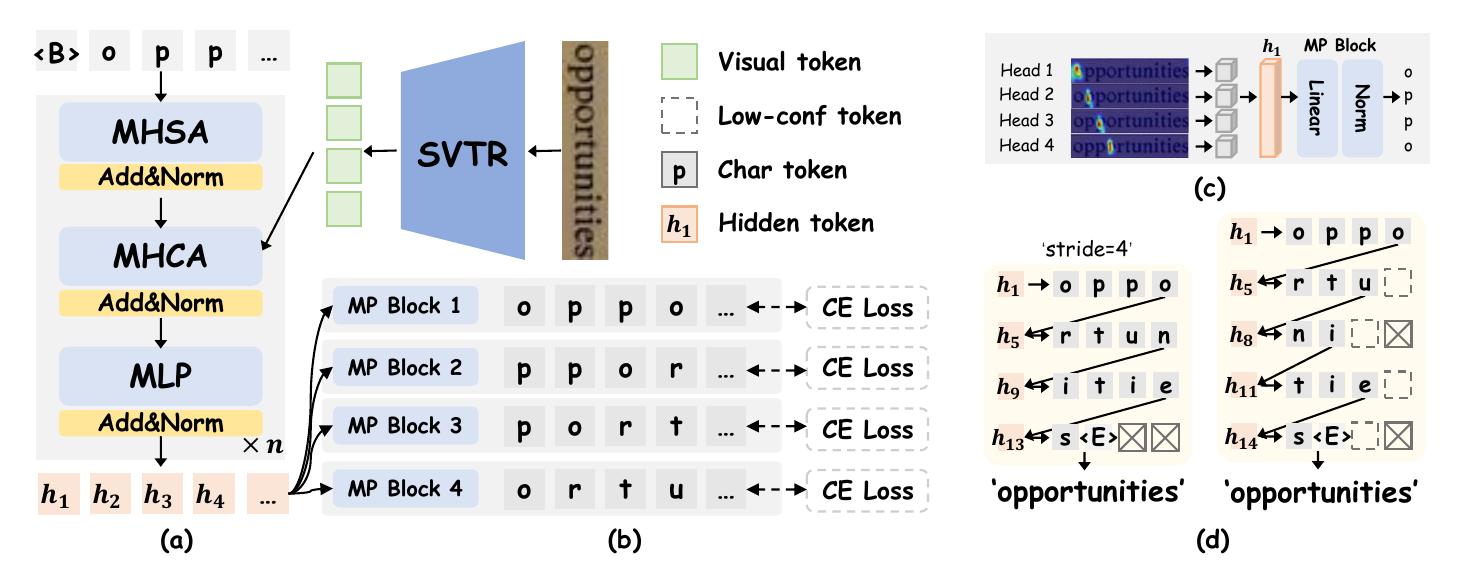}
    \caption{The overview of the proposed \textbf{FlexReader}. (a) The architecture of the Sub-string Perception module. (b) The implementation of Multi-branch Prediction and the loss computation strategy. Each branch independently computes the loss with its corresponding target, and the losses are then aggregated to obtain the final objective. (c) Visualization of the multi-head cross-attention in Sub-string Perception. (d) The decoding procedures of Fixed-step Decoding (left) and Adaptive Prefix Decoding (right).} 
    \label{fig2}
\end{figure*}
\section{Method}
Given a text image $x\in \mathbb{R}^{3 \times H \times W}$, our goal is to recognize the text string $y\in \mathbb{R}^{L_{max}}$ contained in the image.
The proposed FlexReader decomposes decoding into \textit{Sub-string Perception} and \textit{Multi-branch Prediction}, enabling variable-length decoding within a single model. We further introduce \textit{Adaptive Prefix Decoding}, which adaptively selects the decoding length at each step based on the model's outputs, thereby identifying the optimal decoding path. An overview of our method is shown in Fig.~\ref{fig2}.

\subsection{Vision Encoder}
We employ SVTR~\cite{ijcai22/svtr/YKDu}, a widely used visual encoder in the STR field, for feature extraction. Given an image $x$, the feature extractor processes the input to obtain the visual feature:
\begin{equation}
F = \operatorname{\textit{SVTR}}(x) \in \mathbb{R}^{C \times h \times w},
\end{equation}
where $C$ denotes the feature dimension, and $h$ and $w$ denote the spatial dimensions of the feature map.

\subsection{Sub-string Perception}
\label{sec:substring_perception}
\textit{Sub-string Perception (SP)} takes the previously decoded information as input, perceives the entire future sub-string, and encodes it into a single token. Specifically, at each decoding step $i$, self-attention is applied to encode the previously decoded context $E_{0:i-1}$:
\begin{equation}
H_{0:i-1} = \operatorname{\textit{Self-Attn}}(E_{0:i-1} + PE_{0:i-1}) \in \mathbb{R}^{C \times i},
\end{equation}
where $E_{0:i-1}$ denotes the textual embeddings of the decoded characters $y_{0:i-1}$, while $PE_{0:i-1}$ represents the corresponding positional embeddings. Subsequently, we employ causal cross-attention to perceive the sub-string $y_{i+L}$ and encode its information into a hidden state $h_i$:
\begin{equation}
h_{i-1} = \operatorname{\textit{Cross-Attn}}(H_{i-1}, F, M) \in \mathbb{R}^{C \times 1},
\end{equation}
where $h_{i-1}$ represents the hidden state obtained by perceiving the next $L$ characters $y_{i+L}$, and $L$ is the predefined sub-string perception length. $M$ denotes the lower triangular causal mask used to prevent information leakage from future tokens, which is utilized during training to enable the parallel computation of hidden states at all time steps. Fig.~\ref{fig2} (c) visualizes the multi-head cross-attention in the Sub-string Perception (SP) module. As can be observed, different attention heads individually perceive distinct characters within the target substring, and these character-level representations are subsequently concatenated along the feature dimension to form the current hidden state.

\subsection{Multi-branch Prediction}
\label{sec:multibranch_prediction}
\textit{Multi-branch Prediction (MP)} takes the hidden state $h_{i-1}$ at a given time step $i$ and predict the next $L$ characters $y_{i+L}$ independently. We achieve this by employing $L$ parallel decoding blocks. Specifically, given the hidden state $h_{i-1}$, the $n$-th decoding block decodes the $({i+n-1})$-th character from it:
\begin{equation}
y_{i+n-1} = \operatorname{\textit{MPBlock}}_n(h_{i-1}), \quad n=1,\dots,L.
\end{equation}

As discussed above, different characters are encoded in different dimensions of the hidden state. Therefore, each MPBlock only needs to extract the corresponding character representation from the relevant dimensions through a simple linear projection. 

\subsection{Optimization Objective}
We design a dedicated optimization objective to supervise the training process of FlexReader, which consists of two aspects: (1) encouraging the SP module to learn the capability of substring perception; (2) guiding the Multi-branch Prediction module to accurately extract each character from the perceived substring. 
Specifically, each character $y_i$ can be predicted by combining any hidden state in $h_{i-L-1:i-1}$ with its corresponding $MPBlock$ (e.g., $y_{10}$ can be predicted from $h_9$ with $MPBlock_1$, $h_8$ with $MPBlock_2$, and so on). Therefore, we reformulate the optimization objective from the conventional one-to-one paradigm in AR decoding to a one-to-many paradigm, enabling the SP and MP modules to cooperate by optimizing their respective objectives independently:
\begin{equation}
\min_{\theta} \; \mathbb{E}_{(x, y) \sim \mathcal{D}} \left[ - \sum_{t=1}^{L_{max}} \sum_{n=1}^{L} \log p_{\theta}\left( y_{t} \mid y_{<t-n+1}, x \right) \right],
\end{equation}
where $\theta$ denotes all trainable parameters of the model, $L_{\max}$ represents the maximum character length, and $L$ denotes the predefined sub-string perception length.

\subsection{Inference Strategy}
By integrating SP and MP, FlexReader supports Flexble decoding strategies, including fixed-step decoding with different strides and adaptive determination of the decoding length at each step. Fig.~\ref{fig2} (d) provides a visualization of the two decoding strategies.
\paragraph{Fixed-step decoding.} 
Given the predefined substring perception length $L$, the trained model supports fixed-step decoding with arbitrary step sizes in the range of $[1, L]$. Specifically, given a decoding stride $s \leq L$, the model decodes $s$ characters at each decoding step:
\begin{equation}
y_{t+1:t+s} = \operatorname{\textit{Decoder}}(y_{0:t}), \quad n=1,\dots,L_{max}.
\end{equation}

This process continues until the complete sequence is decoded, where the decoding termination is indicated by the $<EOS>$ token. Alg.~\ref{alg:fix_step_decoding} presents the pseudo-code of the proposed algorithm.

As the decoding stride $s$ increases, more characters are decoded at each step, reducing the number of decoding iterations and improving inference efficiency. However, larger strides may slightly compromise contextual reasoning due to less historical information. This flexible decoding scheme allows FlexReader to adopt larger strides for easier test sets to improve efficiency, while using smaller strides for more challenging ones to preserve accuracy.

\paragraph{Adaptive Prefix Decoding}
Previous studies~\cite{mm21/pimnet/ZQiao, ijcv25/ipad/XMYang} have shown that low-confidence predictions are strongly associated with challenging characters lacking sufficient visual evidence, such as those affected by occlusion or blur, which account for a substantial proportion of recognition errors. Therefore, we defer the prediction of these ambiguous characters and allow them to benefit from richer contextual dependencies in subsequent decoding iterations. The procedure of Adaptive Prefix Decoding is presented in Alg.~\ref{alg:adaptive_prefix_decoding}. 

Specifically, given the decoded sequence $y_{0:i}$, we first predict the subsequent sub-string $y_{i+1:i+L}$. We then identify the first character $y_{l}$ with a confidence score below a predefined threshold. Subsequently, we append the selected high-confidence prefix 
$\hat{y}_{i+1:i+l-1}$ to the decoded sequence $y_{0:i}$, while discarding the 
first low-confidence character $y_l$ and its subsequent predictions. The 
updated sequence is then fed back into the decoder for the next decoding 
iteration. The process is repeated until the complete sequence is predicted, e.g., when the <EOS> token is predicted with a confidence score above the predefined threshold.

\begin{algorithm}[t]
\caption{Fixed-step Decoding}
\label{alg:fix_step_decoding}

\KwIn{Visual feature $F_v$, initial token sequence $y_0=\langle BOS\rangle$, step size $s (s\leq L)$}
\KwOut{Predicted text sequence $y$}

Initialize $y\leftarrow[y_0]$, $t\leftarrow0$

\While{$t<L_{\max}$}{

    $\hat{y}_{t+1:t+s}
    =
    \mathrm{Decoder}(y_{\leq t}, F_v)$
    
    \tcp{Update decoded sequence}
    $
    y\leftarrow y\cup\hat{y}_{t+1:t+s}
    $
    
    \If{$\langle EOS\rangle\in\hat{y}_{t+1:t+s}$}{
        \textbf{break}
    }
    $t\leftarrow t+s$

}

\Return{$y$}

\end{algorithm} 

\begin{algorithm}[t]
\caption{Adaptive Prefix Decoding}
\label{alg:adaptive_prefix_decoding}

\KwIn{Visual feature $F_v$, initial token sequence $y_0=\langle BOS\rangle$}
\KwOut{Predicted text sequence $y$}

Initialize $y\leftarrow[y_0]$, $t\leftarrow0$

\While{$t<L_{\max}$}{

    $\left(\hat{y}_{t+1:t+L}, p_{t+1:t+L}\right)
    =
    \mathrm{Decoder}(y_{\leq t}, F_v)$
    
    \tcp{Find the first low-conf prediction}
    $
    l_t=\max\{1, \min\{i\mid p_{t+i}<\tau,\ i\in[1,L]\} - 1\}
    $

    \tcp{Update decoded sequence}
    $
    y\leftarrow y\cup\hat{y}_{t+1:t+l_t}
    $
    
    \If{$\langle EOS\rangle\in\hat{y}_{t+1:t+l_t}$}{
        \textbf{break}
    }
    $t\leftarrow t+l_t$

}

\Return{$y$}

\end{algorithm} 
\section{Experiments}

\subsection{Datasets and Implementation Details}
We train our models on the large-scale real-world Union14M-Filter dataset \cite{arxiv/svtrv2/YKDu}, which contains 3.2 million images covering diverse text scenarios. For evaluation, we employ widely adopted benchmarks in the STR community, including 6 common benchmarks (CoB)~\cite{bmvc12/iiit5k/Mishra, iccv11/svt/KWang, icdar13/ic13/Karatzas, icdar15/ic15/Karatzas, iccv13/svtp/Phan, esa14/cute80/Risnumawan}, 7 Union14M benchmarks (U14M-B)~\cite{iccv23/maerec/QJiang}, covering seven distinct scenarios: curve, multi-oriented, artistic, contextless, salient, multi-words, and general, and 5 more challenging STR benchmarks (MCB), including Uber~\cite{cvprworkshop17/uber/YZhang}, ArT~\cite{icdar19/art/Chng}, COCO-Text~\cite{arxiv/cocotext/Veit}, WordArt~\cite{eccv22/wordart/XDXie}, and OST~\cite{eccv20/robustscanner/XYYue}.

Our models are trained using the AdamW optimizer \cite{iclr19/adamw/Loshchilov} with a weight decay of $0.05$, an initial learning rate of $3.2 \times 10^{-5}$, and a batch size of $512$. We employ a one-cycle learning rate schedule with a linear warm-up phase spanning $1.5$ epochs, and train for a total of $80$ epochs. Input images are uniformly resized to $32 \times 128$, and RandAugment \cite{nips20/randaug/Cubuk} is applied for data augmentation following PARSeq \cite{eccv22/parseq/Bautista}. The maximum text sequence length $L$ is fixed at $25$, while the character vocabulary size $N$ is $96$, covering uppercase and lowercase letters, digits, punctuation marks, and special tokens for sequence start, end, and padding. For the -B and -L variants, the feature dimension $d$ and the number of attention heads $r$ are configured as $(384, 12)$ and $(512, 16)$, respectively. All experiments are performed on a single NVIDIA A100 GPU with 80GB of memory.

\subsection{Results of Different Decoding Strategies}
A key feature of FlexReader is its ability to support multiple decoding strategies within a single model. We therefore evaluate its performance under each strategy. We train two models with sub-string perception lengths of $L=4$ and $L=8$, with the results reported in Table~\ref{tab:decoding_strategy}. The experimental results lead to the following conclusions: 
(1) A model trained with a perception length of $L$ can support decoding with an arbitrary step size $s < L$ during inference, demonstrating the flexibility of FlexReader in the decoding process; (2) As the decoding step size $s$ increases, the recognition accuracy gradually decreases, indicating that decoding with smaller step sizes can more fully exploit the semantic information from previously decoded tokens. However, this also requires more decoding iterations, resulting in a significant increase in decoding time; (3) By leveraging our proposed Adaptive Prefix Decoding, the model can adaptively identify an optimal decoding strategy, thereby achieving a superior speed--accuracy trade-off. A concrete example can be found in the last three rows of the table: with APD, the model achieves recognition accuracy comparable to decoding with $s=1$, while requiring only slightly more inference time than decoding with $s=8$.

\begin{table}[h]
\footnotesize
\centering
\setlength{\tabcolsep}{2.3pt}

\begin{tabular}{c|c|c|cccccccc|c}
\hline

\multirow{2}{*}{\textbf{$L$}}
& \multirow{2}{*}{\textbf{$s$}}
& \multirow{2}{*}{\textit{APD}}
& \multicolumn{8}{c|}{\textbf{Union14M-Benchmarks}}
& \multirow{2}{*}{\shortstack{\textbf{Time}\\\textbf{(ms)}}} \\

\cline{4-11}

& &
& \textbf{CUR}
& \textbf{MO}
& \textbf{ART}
& \textbf{CTL}
& \textbf{SAL}
& \textbf{MTW}
& \textbf{GEN}
& \textbf{Avg}
& \\

\hline

\multirow{4}{*}{4}
& 1 &
& 92.7 & 91.7 & 80.5 & 83.0 & 87.8 & 87.6 & 84.3 & 86.80 & 58.39 \\

& 2 &
& 92.5 & 91.5 & 79.8 & 82.3 & 87.7 & 86.7 & 83.9 & 86.34 & 34.15 \\

& 4 &
& 91.8 & 90.0 & 80.0 & 81.8 & 86.9 & 86.2 & 83.5 & 85.74 & 20.91 \\

& - & $\checkmark$
& 92.6 & 91.1 & 80.3 & 82.8
& 87.5 & 87.8 & 84.3 & 86.63 & 23.97 \\

\hline

\multirow{5}{*}{8}
& 1 &
& 92.5 & 90.8 & 78.1 & 82.1 & 88.1 & 86.2 & 83.8 & 85.94 & 61.18 \\

& 2 &
& 92.0 & 90.8 & 78.0 & 82.0 & 88.1 & 85.1 & 83.4 & 85.63 & 34.98 \\

& 4 &
& 91.6 & 90.6 & 77.6 & 82.0 & 87.6 & 84.3 & 83.1 & 85.26 & 19.37 \\

& 8 &
& 90.8 & 88.6 & 77.8 & 81.1 & 86.9 & 83.5 & 82.7 & 84.49 & 11.71 \\

& - & $\checkmark$
& 92.5 & 90.9 & 77.8 & 81.8 & 88.3 & 85.9 & 83.8 & 85.86 & 13.10 \\

\hline
\end{tabular}

\caption{Comparison under different decoding configurations. $L$ denotes the sub-string perception lengths used during training, while $s$ denotes the decoding step size used during inference.}
\label{tab:decoding_strategy}
\end{table}

\subsection{Comparisons with Existing Decoders}
We conduct experiments to compare our decoding method with several mainstream decoding paradigms across texts of varying lengths, including autoregressive models (ARM), parallel decoding models (PDM), and mask diffusion Models (MDM). Specifically, we categorize all images in Union14M-Benchmarks into three groups according to their text lengths: short ($1-8$), medium ($9-16$), and long ($>16$). We then evaluate the recognition accuracy and decoding time of different decoding methods on each group. For a fair comparison, all decoders employ an identical six-layer Transformer architecture, with the hidden dimension and the number of attention heads consistently set to 384 and 12, respectively. The only differences lie in their training and inference strategies. Moreover, all methods use SVTR as the visual encoder.

\begin{table}[h]
\footnotesize
\centering
\setlength{\tabcolsep}{3pt}

\begin{tabular}{l|cccc|ccc}
\hline

\multirow{2}{*}{\textit{Decoder}}
& \multicolumn{4}{c|}{\textbf{Accuracy}}
& \multicolumn{3}{c}{\textbf{Time (ms)}} \\

\cline{2-8}

& \textbf{Short}
& \textbf{Medium}
& \textbf{Long}
& \textbf{Avg}
& \textbf{Short}
& \textbf{Medium}
& \textbf{Long} \\

\hline

ARM
& \textbf{89.7} & \underline{86.5} & 72.4 & 82.9
& 64.92 & 109.53 & 193.23 \\

PDM
& 87.9 & 81.7 & 67.6 & 79.1
& 6.30 & 6.53 & 7.23 \\



MDM-LC
& 88.9 & 83.6 & 67.6 & 80.1
& 16.88 & 17.59 & 30.74 \\

\hline

FLRE (L=4)$^{\dag}$
& 89.0 & 85.3 & \underline{74.3} & 82.9
& 17.37 & 27.29 & 46.46 \\

FLRE (L=4)$^{*}$
& \underline{89.6} & \textbf{86.9} & 73.3 & \underline{83.3}
& 18.93 & 29.81 & 50.62 \\

FLRE (L=8)$^{\dag}$
& 87.9 & 84.4 & 73.3 & 81.9
& 9.53 & 18.58 & 29.47 \\

FLRE (L=8)$^{*}$
& 89.1 & \underline{86.5} & \textbf{78.1} & \textbf{84.6}
& 11.12 & 20.75 & 32.29 \\

\hline
\end{tabular}

\caption{Comparison of different decoding methods. $^{\dag}$ denotes decoding with the maximum step size, while $^{*}$ denotes decoding with APD. All decoding times are averaged over five repeated measurements on a single A100 GPU.}
\label{tab:decoder}
\end{table}

The experimental results are presented in Table~\ref{tab:decoder}. ARM demonstrates strong recognition performance on short and medium-length texts. However, its token-by-token decoding manner results in significantly longer decoding times than other decoders. Moreover, ARM is prone to error accumulation, leading to inferior accuracy on long texts. Although PDM achieves fast inference, its lack of contextual modeling results in substantially lower recognition accuracy. MDM can be regarded as a compromise between the two, achieving a balance between recognition accuracy and decoding speed. Compared with these methods, our FlexReader demonstrates clear advantages, which can be summarized as follows: (1) On short and medium-length texts, FlexReader achieves recognition accuracy comparable to that of ARM while delivering more than a $3\times$ speedup in decoding. On long texts, FlexReader achieves the best recognition performance; (2) Compared with fixed-step decoding, Adaptive Prefix Decoding achieves significantly better recognition performance with only a marginal increase in inference time, yielding an average improvement of $2.7\%$ when $L=8$.

\begin{figure}[t]
    \centering
    \includegraphics[width=0.5\textwidth]{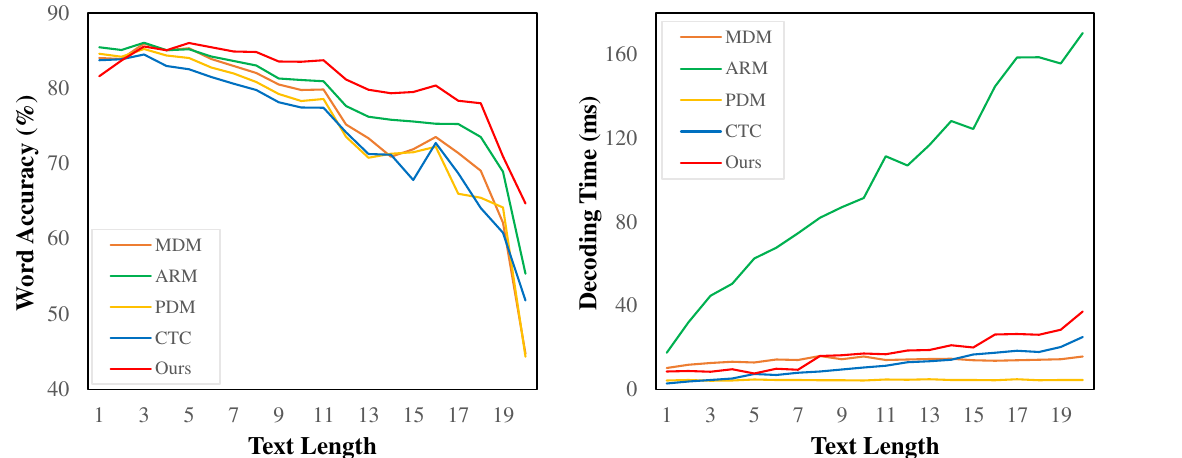}
    \caption{\textbf{Left:} Text length vs. accuracy for different models. 
    \textbf{Right:} Text length vs. decoding time for different models. All decoding times are averaged over five repeated measurements on a single A100 GPU.} 
    \label{fig5}
\end{figure}

To better illustrate the impact of text length on recognition accuracy and decoding efficiency, we group all samples in Union14M-B by text length and evaluate each model on the corresponding subsets. The results are presented in Fig.~\ref{fig5}. As text length increases, all models exhibit a general decline in recognition accuracy, while FlexReader consistently maintains a performance advantage. Notably, this advantage becomes more pronounced for longer text sequences, demonstrating its robustness in long-text recognition. In terms of decoding time, ARM exhibits a sharp increase as text length grows, whereas FlexReader maintains a relatively moderate increase in latency. Although slightly slower than PDM, FlexReader achieves substantially higher recognition accuracy while remaining considerably faster than ARM. These results highlight the effectiveness of FlexReader in balancing recognition accuracy and decoding efficiency.

\subsection{Ablation of the Threshold $\eta$}

\begin{table}[t]
\footnotesize
\centering
\setlength{\tabcolsep}{3pt}

\begin{tabular}{c|cccc|ccc}
\hline

& \multicolumn{4}{c|}{\textbf{Accuracy}}
& \multicolumn{3}{c}{\textbf{Time (ms)}} \\

\cline{2-8}

& \textbf{CoB}
& \textbf{U14M-B}
& \textbf{MCB}
& \textbf{Avg}
& \textbf{CoB}
& \textbf{U14M-B}
& \textbf{MCB} \\

\hline

$\eta=0$
& 96.5 & 87.8 & 83.4 & 89.2
& 7.44 & 8.39 & 8.91 \\

$\eta=0.1$
& 96.5 & 87.8 & 83.4 & 89.2
& 8.15 & 8.49 & 8.97 \\

$\eta=0.3$
& 96.5 & 87.8 & 83.8 & 89.4
& 8.89 & 8.08 & 9.61 \\

$\eta=0.5$
& 96.8 & 88.5 & 84.8 & 90.0
& 8.86 & 9.17 & 10.96 \\

$\eta=0.7$
& \underline{97.0} & \underline{88.8} & \underline{85.3} & \underline{90.4}
& 9.66 & 10.78 & 12.29 \\

$\eta=0.9$
& \textbf{97.1} & \underline{88.8} & \textbf{85.5} & \textbf{90.5}
& 21.79 & 21.42 & 25.44 \\

$\eta=1.0$
& \textbf{97.1} & \textbf{88.9} & \textbf{85.5} & \textbf{90.5}
& 51.01 & 40.59 & 50.32 \\

\hline
\end{tabular}

\caption{Ablation study of the threshold $\eta$. All decoding times are averaged over five repeated measurements on a single A100 GPU.}
\label{tab:eta}
\end{table}

\begin{figure}[t]
    \centering
    \includegraphics[width=0.5\textwidth]{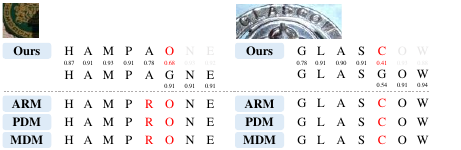}
    \caption{Recognition cases of low-confidence character correction by FlexReader. Red denotes incorrect characters.}  
    \label{fig3}
\end{figure}

We conduct an ablation study on the confidence threshold $\eta$ used in APD. The results are presented in Tab.~\ref{tab:eta}. When $\eta=0$, the model directly accepts all predicted characters at each decoding step, achieving the same performance as Fixed-step Decoding with the maximum receptive length. As $\eta$ increases, the model progressively “revisits” more characters, leading to a gradual improvement in recognition accuracy at the cost of increased decoding time. When $\eta=1$, the model decodes only one character at each step, which is equivalent to Fixed-step Decoding with a step size of 1. Comparing the results across different settings, we find that $\eta=0.7$ provides the most favorable speed–accuracy trade-off. When $\eta<0.7$, the recognition accuracy drops substantially, whereas increasing $\eta$ beyond 0.7 leads to a significant increase in decoding time with only marginal accuracy gains. We therefore set $\eta=0.7$ as the default threshold for APD, allowing characters with confidence scores below 0.7 to be regarded as “highly suspicious” and re-predicted. Fig.~\ref{fig3} shows examples of re-predicting low-confidence characters. The model first predicts all characters within the perceived length. After identifying the first low-confidence character, it preserves the preceding characters and re-decodes the remaining sequence.
 
\subsection{Effectiveness of Sub-string Perception \& Multi-branch Prediction}

\begin{figure*}[t]
    \centering
    \includegraphics[width=1.0\textwidth]{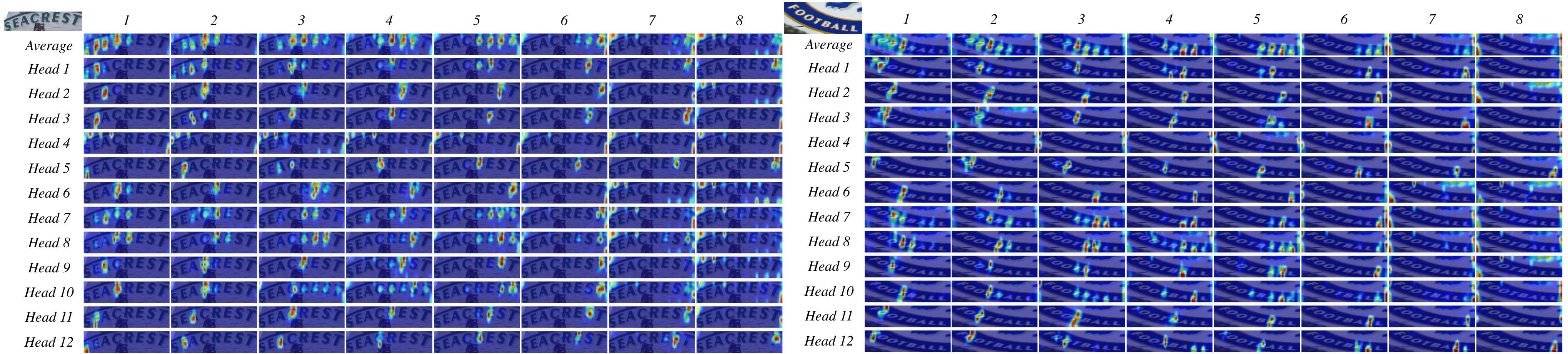}
    \caption{The visualization results of the cross-attention in the final layer. From top to bottom are the average attention across all attention heads and the attention maps of the 12 individual attention heads, respectively; from left to right are the results at each decoding step.} 
    \label{fig4}
\end{figure*}

As described in Sections~\ref{sec:substring_perception} and~\ref{sec:multibranch_prediction}, Sub-string Perception is used to perceive future sub-strings, while Multi-branch Prediction decodes each character in the sub-string separately. We further validate the effectiveness of both components through visualization experiments.

As shown in Figure~\ref{fig4}, we visualize the attention maps of the decoder at each decoding step and compare the attention patterns across different decoding steps, perception lengths $L$, and input images. We observe the following patterns: 
(1) The model is able to perceive the entire sub-string at each decoding step. For the same input image, at different decoding steps $t$, the averaged attention consistently attends to the $t$-th to $(t+L)$-th characters. For example, as shown in the upper-left part of Figure~\ref{fig4}, when $L=4$ and $t=1$, the averaged attention (the \textit{Average} row) focuses on the sub-string $SEAC$, while at $t=2$, it focuses on $EACR$, and so on. Similarly, when $L=8$ and $t=1$, the attention focuses on the sub-string $SEACREST$. At $t=2$, since the length of the remaining sub-string is shorter than 8, the attention focuses on $EACREST$. This demonstrates that Sub-string Perception effectively perceives the sub-string to be decoded at the current step.
(2) Larger perception lengths lead to limited perception of distant characters. As we increase the perception length, although the model is able to perceive longer sub-strings, its ability to capture distant characters becomes weaker. For example, when $L=8$ and $t=1$, the model's perception of the last four characters, $REST$, is only reflected in \textit{head 2} and \textit{head 8}. In contrast to the first four characters, which are consistently captured individually by specific attention heads, the latter four characters are only vaguely perceived as a whole. This explains why, although the model supports decoding with a large step size, its performance still lags behind that of decoding with a smaller step size.
(3) Each attention head learns a fixed target position. For example, in the model with $L=4$, \textit{head 2} consistently focuses on the third character of the sub-string to be decoded: it attends to $A$ at $t=1$, $C$ at $t=2$, and so on. Similarly, \textit{head 3} consistently focuses on the second character of the target sub-string. We observe similar patterns in the model with $L=8$. Interestingly, this pattern is independent of the input image and is determined by the model itself. For example, by comparing the two images shown in Figure~\ref{fig4}, we observe that the same attention head of the same model focuses on the same character position at the same decoding step, regardless of the input image. Note that we do not explicitly supervise individual attention heads during training; therefore, these patterns are spontaneously learned by the model.

These observations explain why the combination of Sub-string Perception and Multi-branch Prediction is effective: Sub-string Perception first receives the already decoded sequence and uses different attention heads to perceive specific characters or groups of characters in the future sub-string, which are then concatenated to form a complete hidden state. Consequently, the information of different characters is encoded into different dimensions of the hidden state. Multi-branch Prediction therefore only needs to learn how to extract and decode the character information from the corresponding dimensions, which explains why it can effectively perform prediction using only multiple independent linear projection layers.

\subsection{Comparison with State-of-the-art}

\begin{table*}[t]\footnotesize
\centering
\caption{Results on English benchmarks tested against existing models when trained on the real-world Union14M-Filter training set. 
\colorbox{bestcolor}{\textbf{Yellow}} and 
\colorbox{secondcolor}{\textbf{Blue}} values denote the 1st and 2nd results in each column. 
$^\dag$ denotes that the result is obtained by training the model on Union14M-Filter using the code they released. $^{\ddagger}$ indicates that we use Union14M-Filter and UnionST as the training sets.
The data presented in the table corresponds to Word Accuracy Ignore Cases (WAIC).}
\setlength{\tabcolsep}{3.2pt}{
\begin{tabular}{c|l|ccccccc|cccccccc|c}
\hline
\multicolumn{2}{c|}{\textbf{\multirow{2}{*}{Method}}} & \multicolumn{7}{c|}{\textbf{Common Benchmarks}}                                                                & \multicolumn{8}{c|}{\textbf{Union14M-L Benchmark}}                                                             & \multirow{2}{*}{\textbf{\begin{tabular}[c]{@{}c@{}}\textit{Params}\\  ($\times 10^6$)\end{tabular}}} \\
\multicolumn{2}{c|}{}                        & \textbf{IC13} & \textbf{SVT}  & \textbf{IIIT} & \textbf{IC15} & \textbf{SVTP} & \textbf{CUTE} & \textbf{Avg} & \textbf{CUR} & \textbf{MTO} & \textbf{ART} & \textbf{CTL} & \textbf{SAL} & \textbf{MTW} & \textbf{GEN} & \textbf{Avg}            &                           \\
\hline
\multirow{3}{*}{CTC}    
& CRNN \cite{tpami17/crnn/BGShi}            
& 91.8 & 83.8 & 90.8 & 71.8 & 70.4 & 80.9 & 81.58 
& 19.4 & 4.5 & 34.2 & 44.0 & 16.7 & 35.7 & 60.4 & 30.70 & 8.3 \\

& SVTR$^\dag$ \cite{ijcai22/svtr/YKDu}           
& 96.7 & 96.7 & 98.2 & 88.6 & 91.3 & 96.5 & 94.70 
& 79.3 & 69.4& 67.1 & 68.8 & 75.8 & 76.8 & 77.0 & 73.48 & 18.1 \\    

& SVTRv2 \cite{arxiv/svtrv2/YKDu}           
& 98.7
& 98.0 
& 99.2
& 91.0 
& 93.6 
& 99.0 
& 96.58 
& 91.0 
& 89.5 
& 78.7 
& 81.3
& 85.9 
& 85.1 
& 82.3 
& 84.91 
& 21.0 \\
\hline
\multirow{10}{*}{PDM}      
& SRN \cite{cvpr20/srn/DLYu}              
& 97.5 & 96.3 & 97.2 & 87.9 & 90.9 & 96.9 & 94.45 
& 78.1 & 63.2 & 66.3 & 65.3 & 71.4 & 58.3 & 76.5 & 68.43 & 51.7 \\

& VisionLAN \cite{iccv21/visionlan/YXWang}   
& 97.1 & 95.8 & 98.2 & 88.6 & 91.2 & 96.2 & 94.50 
& 79.6 & 71.4 & 67.9 & 73.7 & 76.1 & 73.9 & 79.1 & 74.53 & 32.9 \\

& ABINet$^\dag$ \cite{cvpr21/abinet/SCFang}       
& 97.8 & 97.7 & 98.1 & 89.4 & 93.2 & 97.6 & 95.62 
& 82.7 & 77.4 & 68.8 & 69.2 & 77.9 & 72.0 & 77.9 & 75.13 & 36.9 \\    

& MATRN \cite{eccv22/matrn/Na}       
& 97.9 & 98.3 & 98.8 & 90.3 & 95.2 & 97.2 & 96.29 
& 82.2 & 73.0 & 73.4 & 76.9 & 79.4 & 77.4 & 81.0 & 77.62 & 44.3 \\

& MGP-STR \cite{eccv22/mgpstr/PWang}       
& 97.1 & 97.8 & 97.9 & 89.6 & 95.2 & 96.9 & 95.75 
& 85.2 & 83.7 & 72.6 & 75.1 & 79.8 & 71.1 & 83.1 & 78.65 & 148.0 \\

& LPV \cite{ijcai23/lpv/BQZhang}	
& 98.1 & 97.8 & 98.6 & 89.8 & 93.6 & 97.6 & 95.93 
& 86.2 & 78.7 & 75.8 & 80.2 & 82.9 & 81.6 & 82.9 & 81.20 & 30.5 \\

& OTE-PD$^\dag$ \cite{cvpr24/ote/JJXu}	
& 97.1 & 96.3 & 97.9 & 88.7 & 91.8 & 96.2 & 94.65 
& 86.2 & 77.1 & 71.7 & 72.7 & 78.7 & 61.9 & 79.2 & 75.34 & 18.1 \\

& CPPD$^\dag$ \cite{tpami25/cppd/YKDu}	
& 98.2 & 98.1 & 98.8 
& 91.1
& 93.8 & 97.6 & 96.27 
& 87.9 & 81.7 & 75.5 & 76.2 & 83.5 & 81.5 & 81.9 & 81.17 & 26.9 \\

& IGTR-PD$^\dag$ \cite{tpami25/igtr/YKDu}	
& 98.4 & 97.8 & 98.6 & 89.6 & 93.5 & 97.6 & 95.91 
& 88.9 & 91.3 & 75.5 & 76.4 & 83.9 & 78.0 & 82.1 & 82.29 & 24.1 \\

& MDiff4STR-PD~\cite{aaai26/mdiff/YKDu} 
& 98.9
& 98.3 
& 98.9 
& 90.8
& 95.7
& 98.6 
& 96.88
& 92.6 
& 93.6
& 79.0
& 84.7
& 86.7
& 84.5
& 85.7
& 86.69
& 31.9 
\\

\hline
\multirow{16}{*}{ARM}       
& ASTER$^\dag$ \cite{tpami19/aster/BGShi}            
& 96.2 & 94.6 & 97.6 & 87.4 & 88.7 & 93.4 & 92.96 
& 74.2 & 78.8 & 61.3 & 65.9 & 75.9 & 70.6 & 76.9 & 71.97 & 19.1 \\

& NRTR \cite{icdar19/nrtr/FFSheng}             
& 97.8 & 96.8 & 98.1 & 88.9 & 93.3 & 94.4 & 94.89 
& 67.9 & 42.4 & 66.5 & 73.6 & 66.4 & 77.2 & 78.3 & 67.46 & 44.3 \\

& SAR$^\dag$ \cite{aaai19/sar/HLi}             
& 96.7 & 94.0 & 97.7 & 84.8 & 84.5 & 94.4 & 92.03 
& 73.2 & 63.5 & 60.1 & 68.8 & 64.2 & 75.4 & 74.7 & 74.65 & 57.5 \\

& SEED \cite{cvpr20/seed/ZQiao}             
& 94.2 & 93.2 & 96.5 & 87.5 & 88.7 & 93.4 & 92.24 
& 69.1 & 80.9 & 56.9 & 63.9 & 73.4 & 61.3 & 76.5 & 68.87 & 24.0 \\

& RoScanner \cite{eccv20/robustscanner/XYYue}    
& 97.7 & 95.8 & 98.5 & 88.2 & 90.1 & 97.6 & 94.65 
& 79.4 & 68.1 & 70.5 & 79.6 & 71.6 & 82.5 & 80.8 & 76.08 & 48.0 \\

& PARSeq$^\dag$ \cite{eccv22/parseq/Bautista}    
& 97.8 & 97.2 & 98.7 & 90.6 & 94.6 & 96.8 & 95.94 
& 87.6 & 88.3 & 72.3 & 77.4 & 84.0 & 80.8 & 82.6 & 81.93 & 23.8 \\

& MAERec \cite{iccv23/maerec/QJiang}       
& 97.6 & 96.8 & 98.0 & 87.1 & 93.2 & 97.9 & 95.10 
& 81.4 & 71.4 & 72.0 & 80.0 & 78.5 & 82.4 & 82.5 & 78.60 & 35.7 \\

& LISTER$^\dag$ \cite{iccv23/lister/CXCheng}   
& 97.3 & 96.6 & 98.5 & 88.2 & 90.7 & 96.5 & 92.45 
& 87.1 & 88.2 & 72.5 & 78.3 & 79.7 & 81.3 & 80.3 & 81.06 & 20.5 \\

& CDistNet \cite{ijcv24/cdistnet/TLZheng}   
& 97.8 & 98.1 & 98.7 & 89.6 & 93.5 & 96.9 & 95.59 
& 81.7 & 77.1 & 72.6 & 78.2 & 79.9 & 79.7 & 81.1 & 78.62 & 43.3 \\

& BUSNet \cite{aaai24/busnet/JJWei}
& 97.8 & 98.1 & 98.3 & 90.2 & 95.3 & 96.5 & 96.06 
& 83.0 & 82.3 & 70.8 & 77.9 & 78.8 & 71.2 & 82.6 & 78.10 & 32.1 \\

& OTE-AR$^\dag$ \cite{cvpr24/ote/JJXu}   
& 97.7 & 97.8 & 98.2 & 89.6 & 94.4 & 97.9 & 95.95 
& 87.9 & 82.8 & 75.3 & 73.4 & 81.8 & 68.9 & 80.9 & 78.70 & 20.0 \\

& CAM \cite{pr24/cam/MKYang}   
& 96.6 & 96.1 & 98.2 & 89.0 & 93.5 & 96.2 & 94.94 
& 85.4 & 89.0 & 72.0 & 75.4 & 84.0 & 74.8 & 83.1 & 80.52 & 58.7 \\

& SMTR \cite{aaai25/smtr/YKDu}	
& 98.4 & 98.2 & 98.8 & 90.0 & 93.3 & 97.6 & 94.14 
& 90.5 & 
92.6
& 75.5 & 80.0 & 84.9 
& 85.2
& 82.6 & 84.47 & 15.8 \\

& IGTR-AR$^\dag$ \cite{tpami25/igtr/YKDu}	
& 98.6 
& 98.3 
& 98.9 
& 90.4 
& 94.7 
& 97.6 
& 96.42 
& 90.3 
& 92.6
& 77.5 
& 78.8 
& 85.6 
& 81.8 
& 82.9 
& 84.21 
& 24.1 \\

& One2Seq \cite{aaai26/one2seq/ZBMa}	
& 98.5 & 97.7 & 99.0 & 91.0 & 95.0 & \cellcolor{secondcolor}\underline{99.3} & 96.73 
& 91.5 & 
90.8
& 79.7 & 80.4 & 86.1
& 83.4
& 83.5 & 85.06 & 34.7 \\

& MDiff4STR-AR~\cite{aaai26/mdiff/YKDu} 
& \cellcolor{bestcolor}\textbf{99.2}
& 98.1 
& 99.1 
& 91.6 
& 96.0
& 98.6 
& 97.09
& 93.6 
& 94.6
& 81.9
& \cellcolor{bestcolor}\textbf{86.5}
& 88.1
& 89.0
& \cellcolor{secondcolor}\underline{86.7}
& 88.63
& 31.9 
\\

& RISTER~\cite{eccv26/ZBMa/rister} 
& 98.6
& 98.3 
& 98.9 
& 91.1 
& 95.2
& 98.3 
& 96.73
& 93.0 
& \cellcolor{bestcolor}\textbf{96.6}
& 78.2
& 78.6
& \cellcolor{secondcolor}\underline{89.1}
& 86.5 
& 83.0
& 86.42
& 32.8
\\

\hline
\multirow{3}{*}{MDM}    

& PIMNet~\cite{mm21/pimnet/ZQiao}      
& 98.4    
& 97.8    
& 98.9    
& 89.6   
& 94.3     
& 97.9 
& 96.15                                                                      
& 80.3    
& 79.8     
& 68.4      
& 75.9          
& 77.8    
& 68.3                                                     
& 80.8       
& 75.90                                                                        
& 24.2                                                                                                              
\\

& IPAD~\cite{ijcv25/ipad/XMYang}
& 98.1
& 97.7
& 99.0    
& 90.8   
& 95.5   
& 97.9    
& 96.50
& 85.2
& 83.3    
& 72.1    
& 78.4      
& 81.4      
& 73.7    
& 82.2      
& 79.47
& 24.7
\\

& MDiff4STR-LC~\cite{aaai26/mdiff/YKDu} 
& \cellcolor{bestcolor}\textbf{99.2}
& 98.3 
& 99.1 
& 91.6 
& \cellcolor{bestcolor}\textbf{97.1}
& 98.6 
& 97.32
& 93.7 
& 94.4
& 82.0
& \cellcolor{secondcolor}\underline{86.0} 
& 87.9
& 87.7 
& \cellcolor{bestcolor}\textbf{86.8}
& 88.35
& 31.9 
\\

\hline
\multirow{6}{*}{Ours}
& FLRE-SVTR-B
& 98.4 & 98.1 & \cellcolor{secondcolor}\underline{99.3} & \cellcolor{secondcolor}\underline{91.8} & 95.7 & 99.0
& 97.05
& 92.5 & 90.9 & 77.8 & 81.8 & 88.3 & 85.9 & 83.8
& 85.86
& 32.8 \\

& FLRE-SVTR-L
& 98.6 & 98.3 & \cellcolor{secondcolor}\underline{99.3} & 90.8 & 96.3 & \cellcolor{secondcolor}\underline{99.3}
& 97.10
& 93.7 & 93.4 & 79.3 & 82.8 & 89.0 & 86.3 & 84.4
& 86.99
& 65.7 \\

& FLRE-SVTRv2-B
& 98.8 & 98.3 & \cellcolor{secondcolor}\underline{99.3} & 90.8 & \cellcolor{secondcolor}\underline{96.6} & 99.0
& 97.13
& 93.8 & \cellcolor{secondcolor}\underline{95.1} & 80.7 & 81.4 & 88.1 & 87.4 & 84.4
& 87.27
& 33.1 \\

& FLRE-SVTR-B$^{\ddagger}$ 
& \cellcolor{secondcolor}\underline{99.1} & \cellcolor{secondcolor}\underline{98.5} & \cellcolor{secondcolor}\underline{99.3} & 91.4 & \textbf{97.1} & \cellcolor{secondcolor}\underline{99.3}
& \cellcolor{secondcolor}\underline{97.45}
& \cellcolor{secondcolor}\underline{94.0} & 94.3 & \cellcolor{secondcolor}\underline{83.7} & 84.4 & \cellcolor{secondcolor}\cellcolor{bestcolor}\textbf{89.8} & \cellcolor{secondcolor}\underline{90.1} & 83.4
& \cellcolor{secondcolor}\underline{88.53}
& 32.8 \\

& FLRE-SVTR-L$^{\ddagger}$ 
& \cellcolor{secondcolor}\underline{99.1} & \cellcolor{bestcolor}\textbf{98.9} & \cellcolor{bestcolor}\textbf{99.4} & \cellcolor{bestcolor}\textbf{92.4} & \cellcolor{bestcolor}\textbf{97.1} & \cellcolor{bestcolor}\textbf{99.7}
& \cellcolor{bestcolor}\textbf{97.77}
& \cellcolor{bestcolor}\textbf{94.9} & \cellcolor{secondcolor}\underline{95.1} & \cellcolor{bestcolor}\textbf{85.5} & 84.9 & \cellcolor{bestcolor}\textbf{89.8} & \cellcolor{bestcolor}\textbf{90.4} & 84.2
& \cellcolor{bestcolor}\textbf{89.26}
& 65.7 \\
\hline
\end{tabular}}

\label{tab:sota_u14m}
\end{table*}

We compare FlexReader with existing mainstream STR methods, including approaches based on various decoding paradigms, such as CTC, PDMs, ARMs, and MDMs, with the results presented in Table~\ref{tab:sota_u14m}.
Although our primary goal is to achieve flexible and fast inference rather than maximize recognition accuracy, FlexReader still demonstrates competitive performance across multiple datasets. For example, even the most basic FLRE-SVTR-B achieves average accuracies of 97.05\% and 85.86\% on CoB and U14M-B, respectively. Despite its simplicity, it already outperforms most existing STR models. When we scale up the backbone, the resulting FLRE-SVTR-L achieves further improvements in generalization performance. In particular, its recognition accuracy is further improved across multiple benchmarks on U14M-B. In addition, we experimented with using SVTRv2 as the visual encoder. As a result, FLRE-SVTRv2-FLRE achieves further improvements over the original version. In particular, on U14M-B, it improves the average accuracy by 1.41\% over the version using SVTR as the backbone.

Finally, the scarcity of real-world data remains a key factor limiting the performance of STR models. A recent work~\cite{cvpr26/XSYe/unionst} revisits data synthesis as a promising solution. By addressing the lack of diversity in previous synthetic datasets~\cite{arxiv/Jaderberg/mj, cvpr16/Gupta/st}, it once again highlights the importance of synthetic data for STR. Therefore, we further trained our model using the proposed synthetic dataset, \textit{UnionST}, to investigate the effectiveness of these newly developed synthetic data for FlexReader. We report the model performance in the last two rows of the table. The results show that these new synthetic data effectively improve the generalization performance of STR models. In particular, compared with the version trained only on real data, FLRE-SVTR-B improves the average accuracy by 0.40\% on CoB and 2.67\% on U14M-B. The same trend is observed for FLRE-SVTR-L. Ultimately, FLRE-SVTR-L$^{\ddagger}$ outperforms other models of comparable size on 9 out of 13 benchmarks. These results not only highlight the potential of synthetic data for STR, but also demonstrate that FlexReader maintains strong generalization ability while providing flexible and efficient inference. 

\begingroup
\renewcommand{\thefootnote}{}
\footnotetext{Further discussions, additional experimental results, more visualization results and limitations are provided in the Appendix.}
\endgroup

\section{Conclusion}
In this work, we introduce a novel STR decoding framework based on substring perception and multi-branch prediction, enabling a single model to decode with different step sizes. Building upon this framework, we further propose FlexReader, an STR model that adaptively identifies the optimal decoding path. FlexReader overcomes the limitation of conventional STR models that rely on a fixed decoding strategy. Compared with traditional autoregressive decoders, FlexReader achieves up to a 6$\times$ decoding speedup while delivering superior recognition performance across diverse scenarios. 
Owing to its flexibility and efficiency, we believe that FlexReader can serve as an ideal alternative to existing STR decoders.

{
    \small
    \bibliographystyle{ieeenat_fullname}
    \bibliography{main}
}

\clearpage

\end{document}